# Aquila: A Hierarchically Aligned Visual-Language Model for Enhanced Remote Sensing Image Comprehension

Kaixuan Lu[1], Ruiqian Zhang[2], Xiao Huang[3], Yuxing Xie[4]

[1]Piesat Information Technology Co., Lrd., Beijing 100195, China
[2]Chinese Academy of Surveying and Mapping, 100036 Beijing, China
[3]Department of Environmental Sciences, Emory University, Atlanta, GA 30322, USA
[4]University of Science and Technology Beijing, 100083 Beijing, China

**Abstract**

Recently, large vision language models (VLMs) have made significant strides in visual-language capabilities through visual instruction tuning, showing great promise in the field of remote sensing image interpretation. However, existing remote sensing vision language models (RSVLMs) often fall short in capturing the complex characteristics of remote sensing scenes, as they typically rely on low-resolution, single-scale visual features, and simplistic methods to map visual features to language features. In this paper, we present Aquila, an advanced visual-language foundation model designed to enable richer visual feature representation and more precise visual-language feature alignment for remote sensing images. Our approach introduces a learnable Hierarchical Spatial Feature Integration (SFI) module that supports high-resolution image inputs (e.g., 1024×1024) and aggregates multi-scale visual features, allowing for the detailed representation of complex visual information. Additionally, the SFI module is repeatedly integrated into the layers of the large language model (LLM) to achieve deep visual-language feature alignment, without compromising the model's performance in natural language processing tasks. These innovations—capturing detailed visual effects through higher resolution and multi-scale input, and enhancing feature alignment—significantly improve the model's ability to learn from image-text data. We validate the effectiveness of Aquila through extensive quantitative experiments and qualitative analyses, demonstrating its superior performance.

# 1. Instruction

Remote sensing technology, which captures Earth observation data from airborne or satellite platforms, has been widely applied in diverse domains, including agriculture, environmental monitoring, conservation, urban planning, and disaster response. With advancements in artificial intelligence (AI), remote sensing image analysis has become increasingly precise and automated. However, current models predominantly rely on visual inputs, such as those used in target detection, semantic segmentation, and change detection within remote sensing [1-3]. In recent years, the rapid development of large language models (LLMs) has led to significant advancements in natural language processing tasks, enabling human-like dialogue [4-6]. Building on the success of LLMs, vision language models (VLMs) [7-9] have extended the capabilities of traditional visual-only models by incorporating text inputs. These models align visual features with the representation space of language models through visual instruction tuning, facilitating end-to-end integration. VLMs excel in processing multimodal data, demonstrating superior capabilities to earlier models, and are gaining significant attention. Harnessing the potential of VLMs to bridge visual and linguistic data, while developing large-scale models for specific vertical domains, has opened new frontiers in remote sensing applications. This advancement represents a pivotal step in the AI-driven transformation of remote sensing technologies.

In recent years, several notable remote sensing VLMs (RSVLMs) frameworks have emerged within the field of remote sensing. To address the modality gap, some studies have focused on constructing image-text pairs embedded with domain-specific remote sensing knowledge [10-11]. These efforts aim to enhance model comprehension across various remote sensing scenarios, such as image captioning and visual question answering (VQA). Examples include datasets developed for instruction-tuning, such as EarthGPT [12] and RSGPT[13], which contribute to expanding the model's understanding of complex remote sensing tasks. While these initiatives have significantly advanced multimodal capabilities in remote sensing, many of the models rely on vision transformer-based [14] CLIP models to extract visual features, where language supervision plays a dominant role in guiding the vision feature extraction process. For aligning visual and language features, commonly used shallow alignment techniques, such as those employed by InstructBLIP [15] and PaLM-E [16], utilize trainable Q-Formers [17] or linear layers to bridge frozen pretrained vision encoders with language models. These methods map image features into the input embedding space of language models, facilitating the integration of visual data with textual information.

For vision feature extraction strategies, training on high-resolution images offers substantial benefits to VLMs [18], particularly when applied to remote sensing images, which are often rich in features and characterized by complex backgrounds. High-resolution imagery allows these models to detect finer visual details, enabling more accurate identification of individual features, the relationships between them, and the broader context of the scene. The enhanced clarity provided by high-resolution inputs is also critical for generating detailed image descriptions, which are essential in remote sensing applications. To manage the challenge of processing high-resolution images, some methods, such as those implemented in [19], gradually increase image resolution during training, while models like mPLUG-Owl [20] utilize pretrained visual modules designed for larger input resolutions. These techniques enhance the richness of visual embeddings within LLMs by increasing the number of visual tokens derived from high-resolution imagery. However, the computational demands for training on high-resolution images are significant, and challenges remain in effectively processing large-resolution data.

In remote sensing applications, a single feature can exhibit varying characteristics depending on the Ground Sampling Distance (GSD), making multi-scale feature extraction essential for accurately interpreting remote sensing scenes. Despite this, shallow visual-language alignment strategies commonly used in RSVLMs encounter limitations in complex remote sensing environments. These methods, which rely on frozen language model weights, are inherently optimized for processing text tokens and face difficulties when applied to visual data, resulting in a notable misalignment. Even with the availability of detailed remote sensing image-text datasets [21], shallow alignment techniques fall short of achieving deep integration between visual and linguistic data, limiting their effectiveness in fully capturing the intricacies of remote sensing imagery.

In this paper, we present Aquila, a novel method specifically designed to harness the full potential of high-resolution and multi-scale remote sensing images. Just as an eagle captures both vast landscapes and fine details in its gaze, Aquila is designed to achieve a seamless alignment between image and language features, enabling a fluid interpretation that moves effortlessly between overarching contexts and intricate specifics within remote sensing imagery. This unique capability allows Aquila to grasp complex spatial relationships and nuanced scene attributes, delivering a profound, comprehensive understanding of remote sensing environments that parallels the eagle's versatile perspective. To efficiently process high-resolution inputs, we propose the Aquila-CLIP ConvNext (A-CCN) Vision Encoder as the visual backbone, utilizing a convolution-based CLIP architecture (ConvNext CLIP [22]) for enhanced visual encoding.. Compared to vision transformer ViT-based [23] models, the A-CCN Vision Encoder is better suited for handling larger input resolutions and supports multi-scale feature extraction. To effectively integrate high-resolution and multi-scale visual features, we introduce a Hierarchical Spatial Feature Integration (SFI) module, designed to capture detailed visual information across various scales. This module is embedded within our developed Multi-layer Deep Alignment (MDA)-LLM, which performs multiple SFI operations across layers to combine rich visual representations with language instructions, creating a unified sequence input for the LLM. Multiple SFI operations are then applied within the layers of the LLM, combining rich visual representations with language instructions to create a unified sequence input for the LLM. This design enables Aquila to achieve finer-grained semantic understanding, providing detailed interpretations of key image attributes, spatial relationships, and complex scene descriptions.

The main contributions of this work are as follows:

- We introduce Aquila, a novel approach that deeply integrates visual and language features, enabling fine-grained and open-world visual comprehension in remote sensing contexts.
- We develop the SFI module, which leverages a set of learnable query features to fuse high-resolution and multi-scale visual data while preserving the spatial structure of the features. These query features are deeply integrated with language features within the LLM, forming Aquila's unique Multi-layer Deep Alignment (MDA)-LLM, which enhances the model's ability to recognize and interpret complex scenes, particularly those with densely distributed elements.
- We conduct extensive ablation studies to validate the effectiveness of the proposed (SFI) module and the importance of deep visual-language feature fusion.
- We evaluate the Aquila model on multiple datasets, demonstrating its strong performance across a range of tasks, including image caption and general visual question answering.

## 2. Related work

*2.1 Large vision language model*

Building on the strong zero-shot and reasoning capabilities of LLMs [24-30], substantial advancements have been achieved in the development of Large VLMs within the field of computer science. Some VLM studies capitalize on the inherent zero-shot and few-shot learning abilities of pre-trained LLMs to interpret user intent and subsequently invoke external multimodal models. For instance, MiniGPT4 [31] enhances multimodal functionality by integrating a visual module with LLMs, while MM-React[32] introduces a protocol that enables LLMs to decide when to engage visual experts for multimodal tasks. Additionally, Visual ChatGPT[33] implements a prompt manager that instructs ChatGPT on which specialized tools to employ, facilitating the use of 22 different expert models. Lastly, GPT4Tools[34] generates an instruction-following dataset by prompting a high-level teacher model (ChatGPT4) with multimodal contexts, followed by fine-tuning the model using Low-Rank Adaptation (LoRA) [35] to enable expert model invocation.

While these models offer versatile multimodal capabilities, their performance is fundamentally constrained by the limitations of both the LLMs and the external models on which they depend. Another branch of research seeks to improve visual and language processing by aligning the output features of visual encoders with the feature space of language models, thereby enabling end-to-end multimodal integration in VLMs. For instance, Flamingo [36] introduced the Perceiver Resampler, which processes a variable number of input images from the visual encoder and produces a fixed number of visual outputs, enhancing the representation of visual features. BLIP2 [37] employs a trainable Q-Former module to extract a fixed number of output features from the image encoder, effectively bridging the gap between the frozen image encoder and the frozen LLM.

InstructBLIP extends BLIP2 by incorporating instruction fine-tuning, integrating multiple datasets, and assigning weights to data inputs to prioritize task-relevant visual features. The LLaVA series [38-40] enhances image-text understanding by aligning visual encoders with LLMs. Shikra [41], built on the LLaVA framework, specializes in reference dialogue and excels at managing positional input and output.

Although these models have demonstrated impressive multimodal capabilities at the image level, challenges persist in the remote sensing domain, which is characterized by intricate and specialized scenes. As highlighted in [42], the extraction of more detailed image features and the precise alignment between image and language representations remain unresolved issues, emphasizing the need for further advancements in this field.

*2.2 Vision language model for remote sensing*

The application of VLMs in remote sensing is gaining significant momentum, broadly classified into two main categories: contrastive-based VLMs and generative-based VLMs. Contrastive-based VLMs are designed to align representations of similar (positive) image-text pairs while differentiating those of dissimilar (negative) pairs. Notable examples include RemoteCLIP [43] and GeoRSCLIP [44], which leverage large-scale remote sensing image-text datasets and employ contrastive training methods based on the CLIP model, enabling tasks such as scene classification and image-text retrieval.

Conversely, generative-based VLMs focus on learning latent representations that facilitate bidirectional mapping between images and text tokens, allowing the generation of comprehensive image descriptions or detailed textual outputs. For instance, RSGPT [13], the first generative RSVLM in remote sensing, builds upon the InstructBLIP model and utilizes constructed instruction datasets derived from remote sensing scenes to support tasks such as image description and visual question answering. GeoChat [45], trained using the LLaVA-1.5 model with a multimodal instruction-following dataset specific to remote sensing, extends capabilities to address both image-level queries and region-specific interactions, thereby enabling region-focused dialogues. EarthGPT [12] has created the largest multimodal, multi-sensor remote sensing instruction-following dataset, comprising over one million image-text pairs, integrating CNN and ViT architectures to enable robust multimodal visual representations.

Other models, such as SkyEyeGPT [46], SkySenseGPT [47], H2RSVLM [48], and LHRS-Bot [49], build upon the LLaVA architecture and employ highly refined remote sensing-specific instruction datasets to achieve a sophisticated understanding of remote sensing imagery. Despite the significant advancements in multimodal remote sensing achieved by these models, the primary distinction among them lies in the datasets they utilize. Nevertheless, there remains a gap in research focused on enhancing visual representations of remote sensing images and effectively aligning these visual features with corresponding textual descriptions in remote sensing applications.

## 3. Methodology

*3.1 Approach Overview*

The model consists of three fundamental components: the Aquila-CLIP ConvNext (A-CCN) Vision Encoder, the Hierarchical Spatial Feature Integration (SFI) module, and the Multi-layer Deep Alignment (MDA)-LLM, which is based on a pre-trained LLM (Llama-3 [50]) and integrates the SFI module. Figure 1 illustrates the overall structure of the model.

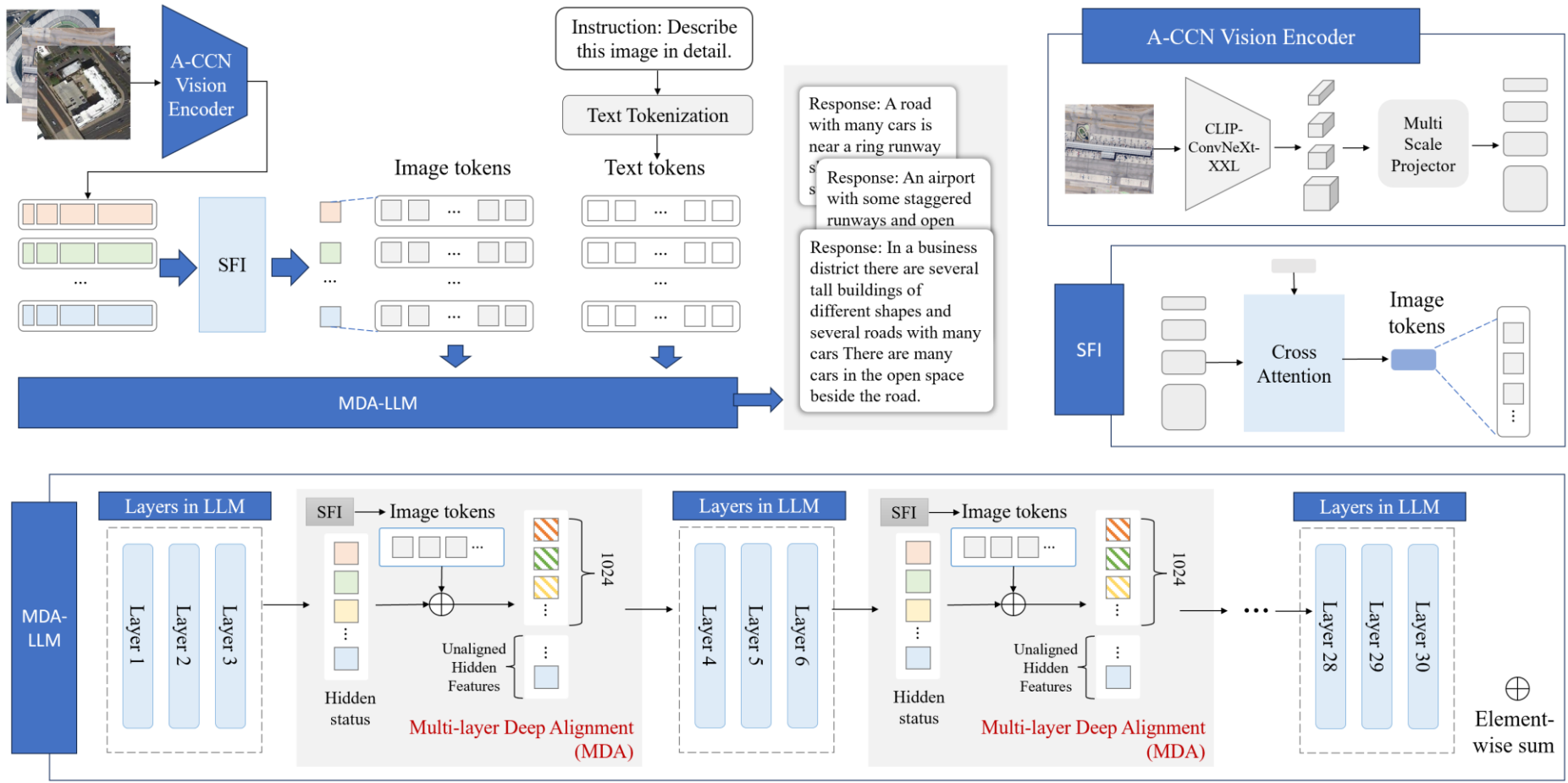

Figure 1. Architecture overview of the proposed Aquila.

## 3.2 Aquila-CLIP ConvNext (A-CCN) Vision Encoder

Most current VLMs utilize the ViT-CLIP model as the vision encoder. However, the input image resolution is typically constrained to lower dimensions, such as 336×336, due to the computational demands imposed by the global attention mechanism in the ViT architecture. This limitation hinders the ability to capture fine-grained details within the image. Furthermore, the ViT architecture lacks the capability to extract multi-scale visual features, which are essential for comprehensive image understanding.

To overcome these challenges, we propose the A-CCN Vision Encoder, which utilizes a convolutional CLIP model, such as CLIP-ConvNext, to support higher-resolution inputs and incorporate multi-scale feature aggregation, following an eagle-eye approach to capture both broad context and fine details. Compared to ViT-based CLIP models, CNN-based architectures not only maintain high performance with higher-resolution inputs but also offer more efficient training and faster inference. Additionally, the multi-scale feature maps generated by the CLIP-ConvNext vision encoder facilitate effective feature aggregation in downstream tasks. In our implementation, we employ the pre-trained CLIP ConvNeXt-XXL model as the vision encoder, which supports image inputs at resolutions up to 1024×1024. In comparison, previous studies only support input image size of 448×448 pixels [51]. We also remove the final layer of the encoder, which is specifically designed for classification-based contrastive learning, to better suit our feature extraction objectives.

Unlike previous methods that only output the final layer of the vision encoder, the A-CCN Vision Encoder outputs multi-scale features from the four stages of CLIP-ConvNeXt. To map these features into the same space as the text features from word embeddings, we use four projectors. Each projector is a two-layer MLP. For each scale of features $f_i \in \mathbb{R}^{C \times H \times W}$, we first reshape it to $f_i \in \mathbb{R}^{(H \times W) \times C}$, where H/W at the four scales are [256, 128, 64, 32] and C at the four scales are [384, 768, 1536, 3072]. The input dimensions of the four projectors correspond to the feature dimensions at each scale ([384, 768, 1536, 3072]), while the output dimension is unified to 1024.

*3.3 Hierarchical Spatial Feature Integration (SFI)*

In remote sensing imagery, the same object may exhibit varying characteristics at different GSDs, underscoring the importance of multi-scale features for comprehensively interpreting remote sensing scenes. We investigate the potential of multi-scale features derived from a vision encoder to leverage their distinctive representations, thereby enhancing the performance of Large VLM. A common approach involves interpolating multi-scale features to a uniform size and concatenating them along the channel dimension. While effective, this method can result in information loss, particularly when high-resolution features are downsampled to lower resolutions. To address this limitation and fully exploit multi-scale feature information, we propose a more advanced strategy, i.e., SFI, which utilizes cross-attention mechanisms and introduces a set of learnable query features that effectively aggregate information from multi-scale features while preserving the spatial structure unique to each scale.

To capture cross-contextual information from multi-scale features, we initialize a learnable feature as the query feature, $X_q \in \mathbb{R}^{L^2 \times C}$, with multi-scale visual features serving as the key and value features, $X_s \in \mathbb{R}^{L_s^2 \times C}$, where $L$ represents the height/width of the learnable query feature. When flattened, it forms the query feature token length $L^2$, and $C$ is the hidden dimension. To ensure consistency across spatial regions, we maintain a proportional relationship between $L$ and $L_s$ (e.g., the size of the $i$-th scale feature is $k_s L \times k_s L \times C$, where $k_s$ is a positive integer). Since the multi-scale features from the encoder are well-aligned spatially, we aim to maintain spatial structure and learn more spatially-aware features during the cross-attention process. Each token in the query is aligned with a specific sub-region in the multi-scale feature maps, meaning cross-attention is not applied globally, but within these sub-regions. Specifically, for each token in the query feature $X_q$, attention is computed only within the corresponding mapped region in the multi-scale feature maps. After computing cross-attention for each token, the results are merged into a final $X_q$, which contains both global information and fine-grained sub-region details from the multi-scale features. The equations are as follows:

$$X_i^{Q^*} = X_i^Q + Softmax(\frac{X_i^Q \cdot X_i^K}{\sqrt{C}})X_i^V$$

$$X_i^Q = W^Q \cdot X_q(i)$$

$$X_i^K = [W^{K_1} \cdot X_1(i_1), W^{K_2} \cdot X_2(i_2), \ldots, W^{K_s} \cdot X_s(i_s)]^T$$

$$X_i^V = [W^{V_1} \cdot X_1(i_1), W^{V_2} \cdot X_2(i_2), \ldots, W^{V_s} \cdot X_s(i_s)]$$

$$X_q = Concat(X_1^{Q^*}, X_2^{Q^*}, \ldots, X_{L^2}^{Q^*})$$

where $X_q(i)$ is the token feature vector at the position $i$ of the query feature, calculated using the query projection matrix $W^Q \in \mathbb{R}^{C \times C}$. $X_s(i_s)$ is the token feature vector at position $i_s$ of the $s$-th scale feature, calculated using the key and value projection matrices $W^{K_s} \in \mathbb{R}^{C \times C}$ and $W^{V_s} \in \mathbb{R}^{C \times C}$, respectively. The position $i_s$ represents a range of positions corresponding to the mapped area of each scale feature relative to the query position $i$.

In addition, recognizing the importance of positional information for fine-grained image understanding, we employ absolute positional encodings in the cross-attention mechanism to minimize the potential loss of positional details during compression. By aggregating sub-region features from multiple scales into a single token, and concatenating all tokens, the resulting feature representation effectively expresses the image information while significantly reducing the number of tokens. This is particularly important for training under limited hardware resources.

*3.4 Multi-layer Deep Alignment (MDA)*

As illustrated at the bottom of Figure 1, we utilize the vision encoder and SFI to obtain image-level embeddings. For the textual data, we tokenize the text sequence using the pre-trained LLM (Llama-3) tokenizer and project the tokens into text embeddings. Given that the SFI model represents complex and abundant image semantics from multi-scale features using a single token, when both the visual and text tokens are input into the LLM, the visual information may degrade as the number of decoder layers increases. To mitigate this issue, we introduce a Multi-layer Deep Alignment (MDA) strategy, incorporating multiple SFI operations within the LLM layers to form the innovative MDA-LLM architecture. This enables the LLM to retain access to uncompressed visual data across layers, preserving the depth and integrity of visual information and ensuring robust visual-textual representation throughout the decoding process.

*3.5 Training*

Drawing inspiration from LlaVA, our model implements a two-stage training framework consisting of pre-training and instruction fine-tuning. Both phases leverage next-token prediction loss for supervised learning, wherein the model predicts the subsequent token in a given input text sequence. The initial stage focuses on aligning image features with word embeddings using simple image-text pairs, while the second stage refines the model's capacity to process complex scenarios through instruction fine-tuning.

- Stage 1 (Image-Text Alignment Pre-training): In the first stage, we initialize the A-CCN vision encoder with pre-trained CLIP-ConvNeXt-XXL weights and the LLM with pre-trained Llama-3 weights. During this stage, only the SFI module parameters remain trainable, while the weights of both the vision encoder and the LLM are frozen. This ensures that the image features are aligned with the embedding space of the language model without altering the pre-trained representations of the LLM itself.

- Stage 2 (End-to-End Fine-Tuning): Following the alignment achieved in Stage 1, the model is capable of interpreting image features. To further enhance its ability to follow instructions and manage complex dialogue scenarios, we engage in a second stage of fine-tuning with instruction-based data. In this stage, the pre-trained SFI module from Stage 1 is used, and while the vision encoder remains frozen, the SFI parameters are trainable. Additionally, we employ the LoRA method to fine-tune the LLM, optimizing the model's adaptability to instruction-following tasks with improved efficiency.

## 4. Experiments

To verify the superior performance and generalization of Aquila, we evaluate our model on a series of vision-language benchmarks. These benchmarks cover a comprehensive range in the multimodal remote sensing domain, primarily including tasks such as image captioning and visual question answering.

### *4.1 Implement Details*

#### *4.1.1 Datasets*

The image-text data utilized for training are entirely publicly available and comprise two primary components. The first component is used for alignment pre-training and includes datasets such as CapERA [52], UCM [53], Sydney [53], NWPU [54], RSICD [55], RSITMD [56], RSVQA-HR [57], RSVQA-LR [57], and WHU_RS19 [58]. We integrated these datasets following the LLaVA data format, resulting in approximately 1 million image-text pairs. The second component is employed for instruction fine-tuning, we leverage FIT-RS [47], a dataset specifically constructed for complex comprehension tasks, containing up to 1.8 million high-quality instruction image-text pairs.

Given that these two datasets encompass instructions, questions, and answers of diverse types and formats (e.g., brief answers consisting of single words or standard answers formulated as complete sentences), it is crucial to specify the required answer format within the instruction prompts. To mitigate ambiguity and potential conflicts, we incorporate prompts such as "Answer the question using a single word or phrase." after questions where the expected answer is a word or phrase. By augmenting the data with these system prompts, we are able to train a cost-effective multimodal model capable of understanding and executing different types of tasks.

*4.1.2 Summary of the Evaluation Benchmarks*

This section provides a comprehensive overview of the evaluation benchmarks employed in our experiments, along with their corresponding metrics (Table 1). For the image captioning task, we selected four widely used datasets:

- RSICD [55]: A dataset specifically designed for remote sensing image captioning tasks. The images are standardized to 224×224 pixels and cover various resolutions.

- Sydney [53] and UCM [53]: Datasets utilized for high-resolution remote sensing image descriptions, providing detailed imagery for more precise captioning.

- FIT-RSFG-Captions [47]: A dataset dedicated to fine-grained remote sensing image descriptions, focusing on detailed and nuanced captioning of remote sensing imagery. For the VQA task, we chose three datasets:

- RSVQA-LR [57]: A VQA dataset tailored for low-resolution remote sensing images, testing the model's ability to infer information from less detailed visuals.

- RSVQA-HR [57]: A VQA dataset focusing on high-resolution remote sensing images, challenging the model to process and understand detailed imagery.
- FIT-RSFG-VQA [47]: A dataset concentrating on fine-grained VQA in remote sensing, emphasizing the model's capacity for detailed analysis and comprehension.

Table 1. Summary of the evaluation benchmarks.

| Dataset | Task | Split | Metric |
| --- | --- | --- | --- |
| RSICD [55] | Image caption | test | Bleu1 |
| Sydney [53] | Image caption | test | Bleu1 |
| UCM [53] | Image caption | test | Bleu1 |
| FIT-RSFG-Captions [47] | Image caption | val | Bleu1 |
| RSVQA-LR [57] | Visual question answering | test | Accuracy |
| RSVQA-HR [57] | Visual question answering | test | Accuracy |
| FIT-RSFG-VQA [47] | Visual question answering | val | Accuracy |

*4.1.3 Training*

Our training methodology employs pre-trained weights from a vision encoder (CLIP-ConvNeXt) and a large language model (Llama3) and consists of two stages. In the first stage, we train on a dataset of 1 million image-text pairs for one epoch, freezing both the vision encoder and the LLM. In the second stage, we fine-tune the LLM using the Low-Rank Adaptation (LoRA) method on a dataset of 1.8 million instruction image-text pairs for one epoch, while keeping the vision encoder frozen. To ensure high image resolution, input images are resized to 1024×1024 pixels in both stages. We use the AdamW optimizer and adopt a cosine annealing scheduler for the learning rate. All training is conducted with a global batch size of 32 on four NVIDIA A800 GPUs. The settings and hyperparameters applied in alignment pretraining and instruction finetuning are listed in Table 2.

Table 2. Hyperparameters for alignment pretraining and instruction finetuning.

| **Hyperparameters** | **Alignment pretraining** | **Instruction finetuning** |
| --- | --- | --- |
| Visual encoder | CLIP-ConvNeXt-XXL | CLIP-ConvNeXt-XXL |
| Textual encoder | Llama-3-8b | Llama-3-8b |
| Datasets | 1M | 1.8M |
| Epoch | 1 | 1 |
| GPUs | 4 NVIDIA A800 | 4 NVIDIA A800 |
| Training time | 70h | 120h |
| Input resolution | $1024^2$ | $1024^2$ |
| Image token length | 1024 | 1024 |
| Global batch size | 32 | 32 |
| Learning rate | 1e-3 | 4e-5 |
| Learning rate scheduler | Cosine | Cosine |
| Optimizer | AdamW | AdamW |
| Adam $\beta$ | (0.9, 0.95) | (0.9, 0.95) |
| Weight decay | 0.05 | 0.1 |
| Warmup ratio | 0.06 | 0.03 |
| Lora rank | -- | 64 |
| Lora alpha | -- | 16 |
| Lora drop | -- | 0.05 |

## *4.2 Results*

We present our results on tasks including image captioning and visual question answering (VQA). For image captioning, we employ the BiLingual Evaluation Understudy (BLEU) [59] metric with unigram precision (BLEU-1). For VQA, we report the average accuracy across all questions.

*4.2.1 Image Captioning.*

This task involves the automatic generation of natural language descriptions of image content, requiring the model to both comprehend the image and articulate its content fluently. We assessed our model's image captioning capabilities using the RSICD, Sydney Caption, RSITMD, and FIT_RSFG-Captions benchmarks [47]. As shown in Table 3, our model Aquila achieves superior performance on three simple description datasets, exceeding RSGPT by 4.28%, 1.16%, and 2.13% in the BLEU-1 metric, respectively. For the complex description dataset FIT_RSFG-Captions, we outperform the state-of-the-art method SkySenseGPT by 7.77%.

Table 3. Performance of VLMs on remote sensing caption datasets

| Model | RSICD | Sydney | UCM | FIT_RSFG-Captions |
|---|---|---|---|---|
| LLaVA1.5-7B | 55.26 | 52.21 | 57.58 | 15.38 |
| GeoChat | 45.86 | 39.79 | 44.09 | 8.79 |
| RSGPT | 70.32 | 82.26 | 86.12 | -- |
| SkySenseGPT | -- | -- | -- | 27.31 |
| Aquila | **74.6** | **83.42** | **88.25** | **35.08** |

*4.2.2 Visual Question Answering (VQA).*

VQA tasks require models to generate correct answers based on input images and accompanying questions, necessitating not only a thorough understanding of the image content but also the ability to infer accurate answers from visual cues in response to specific queries. We evaluated our model's performance on three benchmarks: RSVQA-LR, RSVQA-HR, and FIT_RSFG-VQA [47]. As presented in Table 4, our model Aquila demonstrates significant improvements over the nearest competing methods, surpassing them by an average margin of 1.41%. These results underscore the effectiveness of our approach.

Table 4. Performance of VLMs on remote sensing visual question answer datasets

| Model | RSVQA-LR | RSVQA-HR | FIT_RSFG-VQA |
|---|---|---|---|
| LLaVA1.5-7B | 65.86 | 53.99 | 58.59 |
| GeoChat | 91.81 | 70.82 | 53.47 |
| RSGPT | 92.29 | 92.15 | -- |
| LHRS-Bot | 89.19 | 92.55 | -- |
| H$^2$RSVLM | 89.12 | 74.35 | -- |
| SkyEyeGPT | 88.23 | 86.87 | -- |
| SkySenseGPT | 92.69 | 82.56 | 79.76 |
| Aquila | **92.72** | **92.64** | **83.87** |

## 4.3 Ablation Study

To validate the effectiveness of the key modules in our model design, we conducted the following ablation experiments.

### 4.3.1 SFI vs. Concat

We employed the SFI module to perform multi-scale feature aggregation. The aggregated features encompass both the global information of multi-scale features and the refined details of their sub-regions. To assess the effectiveness of SFI, we replaced it with a concatenation module (Concat) as the baseline. Specifically, we constructed a convolutional module to aggregate the four multi-scale features into a single feature, serving as the final output of the visual encoder. As shown in Table 5, without performing deep alignment, SFI outperforms Concat in image captioning and VQA tasks by 5.62% and 6.85%, respectively. These experimental results indicate that the SFI module, based on a cross-attention mechanism corresponding to local regions, achieves superior feature fusion.

*4.3.1 Multi-layer Deep Alignment(MDA)*

The deep align module facilitates comprehensive alignment between visual and textual features. To evaluate its impact on model performance, we conducted an ablation study by removing the deep alignment module and directly inputting the visual and textual features into the LLM. As shown in Table 5, the absence of deep alignment resulted in performance decreases of 2.55% and 4.64% in image captioning and VQA tasks, respectively, demonstrating the effectiveness of deeply aligning visual and textual features. Furthermore, the combined use of the SFI module and MDA yielded the best results, confirming the efficacy of aggregating and aligning multi-scale visual information.

Table 5. Ablation study on SFI and MDA for Aquila.

| Method | FIT_RSFG Captions | FIT_RSFG VQA |
|---|---|---|
| Baseline | 26.67 | 75.29 |
| +MDA | 29.22 | 79.93 |
| +**SFI** | 32.29 | 82.14 |
| +**SFI** +MDA | **35.08** | **83.87** |

*4.4 Demonstrations*

The demonstrations provided in Figures 2, 3, and 4 offer some insights into Aquila's capabilities in remote sensing image comprehension and VQA. Through these examples, we can highlight Aquila's distinct advantages over existing models, such as GPT-4V, especially when interpreting high-resolution, multi-scale remote sensing imagery.

One of the key advantages of Aquila lies in its SFI module, which allows the model to extract and integrate multi-scale visual features efficiently. This capability is crucial in remote sensing applications, where a single feature can exhibit different characteristics depending on the GSD or spatial resolution. Aquila's SFI module, through a cross-attention mechanism, enables the model to retain the spatial structure and details unique to each scale, preserving fine-grained visual information without compromising the broader context. Unlike GPT-4V, which primarily relies on single-scale processing, Aquila's multi-scale approach ensures that high-resolution images are processed with greater contextual sensitivity. This results in more accurate and contextually aware descriptions of objects and their spatial relationships, as evidenced in Figure 2, where Aquila can precisely describe complex scenes, such as airports, with details of runways, planes, and orientation.

Another advantage is Aquila's MDA strategy, which ensures that the model maintains a robust alignment between visual and textual features across multiple layers. In Figure 3, for instance, we see that Aquila provides a more nuanced and layered description of the scene compared to GPT-4V, which often loses spatial details due to shallow alignment techniques. Aquila's high-resolution image input capability (up to 1024×1024 pixels) also sets it apart from other models. Most models, including GPT-4V, struggle with high-resolution imagery due to computational limitations. This higher resolution input capability enhances Aquila's ability to detect finer visual details that are often critical in remote sensing tasks, such as distinguishing small objects or identifying subtle spatial patterns. Figure 4 demonstrates this advantage, as Aquila accurately answers questions about object positions, colors, and orientations, a level of detail that GPT-4V often misses due to its lower-resolution constraints.

Furthermore, Aquila is specifically designed for remote sensing applications, with training on remote sensing-specific image-text pairs. This specialization allows Aquila to understand domain-specific terminology and spatial configurations unique to remote sensing, such as docks, harbors, and urban layouts, as shown in the qualitative results.

GPT-4V, a more generalized model, lacks this tailored understanding, often resulting in less accurate descriptions and answers when applied to complex remote sensing images. For instance, Aquila's responses include precise descriptions of infrastructural elements and environmental features, which are frequently misinterpreted or oversimplified by GPT-4V due to its broader, non-specialized training.

| Visual input examples | 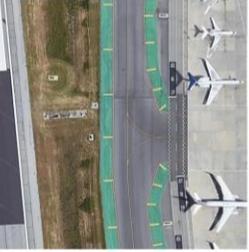 | 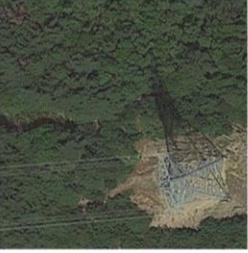 |
|---|---|---|
| User 🦉 | How many planes do this image contain?<br><br>4 | Where is the power tower located in the image?<br><br>A fully visible large lattice tower is on the bottom right part of the image. |
| User 🦉 | Describe the scene.<br><br>The scene depicts an airport, with a grassy area and a runway in the background. Multiple airplanes can be seen in the image. One airplane is located in the lower right corner, partially visible, with another airplane parked beside it. Facing to the right, a third airplane is parked next to the second airplane. Additionally, a fourth airplane is situated in the upper right corner of the image. These airplanes are neatly lined up in a row, each parked alongside another airplane. The airplanes are arranged neatly, creating a sense of order in the scene. | Describe the scene.<br><br>In the forest scene, there is a lattice tower standing tall among the trees. The lattice tower is located towards the right side of the image, slightly closer to the foreground. It is positioned near the edge of the forest, overlooking the trees and the power line that runs through the area. The lattice tower is fully visible in the image, with all its intricate details clearly visible. The power line extends from the lattice tower, disappearing into the distance. The tower stands out against the backdrop of the dense forest, adding a man-made element to the natural landscape. |
| User 🦉 | What colors are on this airplane?<br><br>The airplanes are white in color. | What is the direction of the power lines in the image?<br><br>The power lines appear to extend from the bottom left to the top right. |

Figure 2. VQA demonstrations of the proposal Aquila.

| Visual input examples | 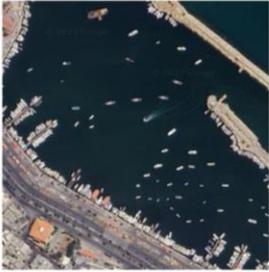 | 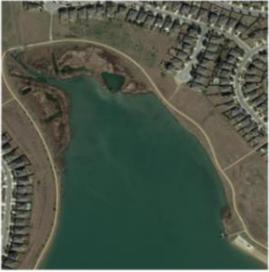 |
|---|---|---|
| User | Describe this image. | Describe this image. |
|  | The scene depicts a harbor. The scene features numerous boats and yachts docked along the waterfront, as well as some vessels scattered throughout the water. There are also breakwaters or piers that help enclose the harbor, providing protection from waves. The area adjacent to the water appears to be a developed urban space with roads, buildings, and possibly parking facilities. The overall setting suggests a coastal city or town with a busy marina for recreational or commercial boating activities. | The image shows a residential area adjacent to a body of water, likely a lake or pond. The neighborhood consists of houses arranged along curving streets, with some homes facing the waterfront. There is a paved pathway that follows the shoreline, creating a walking or biking trail around the lake. |
| 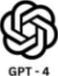 | This image captures an aerial view of a busy maritime area, possibly a harbor or marina. It shows numerous boats, both small and larger vessels, scattered throughout the water, indicating a high level of activity. There are two jetties extending into the water, lined with more moored boats. Along the shore, there is a densely packed urban environment with buildings close to the waterfront. This scene suggests a coastal city with significant boating activity, possibly for both recreational and commercial purposes. The layout and density of the boats hint at a well-used and possibly tourist-friendly port area. | This image shows an aerial view of a residential area surrounding a lake. The lake, which is a turquoise color, is irregular in shape with several peninsulas and inlets. The surrounding area includes multiple single-family homes, arranged in a curvilinear street pattern that follows the contour of the lake. The homes are closely spaced, each with its own backyard, some of which extend almost to the water's edge. There are visible roads and a small dock or pier on the near side of the lake. The area looks well-developed and possibly suburban, with open spaces and possibly recreational areas near the water. |

Figure 3. Comparing the performance of the proposed Aquila with GPT-4V in terms of image describing.

| Visual input examples | 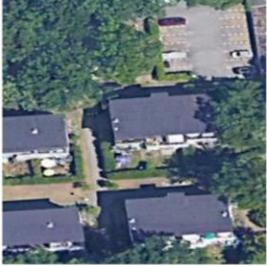 | 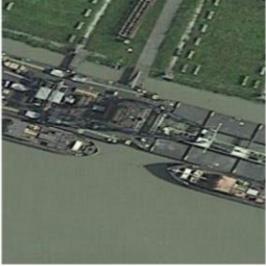 |
|---|---|---|
| User | Is there a car in the image? | How many ships in the image? |
| Aquila | Yes, there are 2 cars in the image. | 2. |
| GPT-4 | Yes, there is 1 car in the image. | In the image, there are two ships visible. |
| User | Is there a ship in the image? | Is there a car in the image? |
| Aquila | No. | No. |
| GPT-4 | No, there is no ship in the image. | No, there are no cars visible in the image. |
| User | What is the color of the roofs? | What is the location of the dock? |
| Aquila | The roofs in the image are black. | The dock is located in the middle of the image, with two ships docked beside it. |
| GPT-4 | The roofs in the image are primarily dark in color, possibly shades of gray or black. | The dock in the image is located along the right side of the frame, extending horizontally into the body of water. |

Figure 4. Comparing the performance of the proposed Aquila with GPT-4V in terms of simple question and answering.

## 5. Conclusion

In this paper, we introduced Aquila, an advanced visual-language model tailored to enhance remote sensing image understanding by effectively integrating high-resolution and multi-scale visual features with language representations. Addressing the limitations of existing RSVLMs—which often rely on low-resolution inputs and simplistic visual-language mappings—Aquila leverages a novel SFI module that supports high-resolution images up to 1024×1024 pixels and aggregates multi-scale visual data, capturing fine-grained details essential for interpreting complex remote sensing scenes.

To bridge the gap between visual and linguistic modalities, we developed an MDA strategy that repeatedly integrates the SFI module within LLM layers. This deep alignment ensures a more precise and robust fusion of visual and language features without compromising the LLM's performance in natural language processing tasks. Through extensive experiments on various benchmarks—including image captioning and VQA—Aquila demonstrated superior performance compared to state-of-the-art models. Our ablation studies further validated the effectiveness of the SFI module and the importance of deep visual-language feature fusion. The collaborative synergy of the SFI and MDA modules enables Aquila to achieve a finer-grained semantic understanding and provides detailed interpretations of key image attributes, spatial relationships, and complex scene descriptions. Notably, when compared to large multimodal models like GPT-4V, Aquila exhibits competitive performance, particularly in focusing on textual information and capturing intricate image details.

In general, the proposed Aquila represents a significant advancement in AI-driven remote sensing technologies, paving the way for more accurate, detailed, and context-sensitive analyses of remote sensing data. Future work may explore optimizing computational efficiency and extending Aquila's capabilities to other remote sensing tasks, thereby fostering more comprehensive solutions in various application domains.

**Acknowledgment**

Acknowledgments will be provided upon acceptance of the article.

**Data Availability**

Data availability will be provided upon acceptance of the article to secure double blind review.